\RequirePackage{fix-cm}
\documentclass{svjour3}                     
\smartqed  
\usepackage{graphicx}

\usepackage{amsmath}
\usepackage{amsfonts}
\usepackage{amssymb}
\usepackage{makeidx}
\usepackage{graphicx}
\usepackage{footnote}
\usepackage{subfigure}
\usepackage{multirow}
\usepackage{hhline}
\usepackage{hyperref}
\usepackage{textcomp}
\usepackage[usenames]{color}
\usepackage[table]{xcolor}

\usepackage{verbatim}
\begin{document}

\title{Application of a Convolutional Neural Network for image classification to the 
analysis of collisions in High Energy Physics}

\titlerunning{CNN for analysis of collisions in HEP }        

\author{Celia Fern\'andez Madrazo \and Ignacio Heredia Cacha \and Lara Lloret Iglesias 
       \and Jes\'us Marco de Lucas
}


\institute{  Instituto de Fisica de Cantabria, IFCA (CSIC-UC) \\
              Tel.: +34 942201458\\
              Fax: +34 942200935\\
              \email{marco@ifca.unican.es}
}



\maketitle

\begin{abstract}
The application of deep learning techniques using convolutional neural networks to the 
classification of particle collisions in High Energy Physics is explored.
An intuitive approach to transform physical variables, like momenta of particles and jets,
into a single image that captures the relevant information, is proposed.
The idea is tested using a well known deep learning framework on a simulation dataset, including leptonic ttbar events and the corresponding background at 7 TeV from the CMS experiment at LHC, available as Open Data. This initial test shows competitive results when compared to more classical approaches, like those using feedforward neural networks. 

\end{abstract}

\section{Introduction}
\label{intro}

Deep learning with convolutional neural networks (CNNs) has revolutionized the world of computer vision and speech recognition over the last few years, yielding unprecedented performance in many machine learning tasks and opening a wide range of possibilities \cite{LeCun2015}.  

 In this paper, we explore a particular application of CNNs, image classification, in the context of analysis in experimental High Energy Physics (HEP). Many studies in this field, including the search for new particles, require solving difficult signal-versus-background classification problems, hence machine learning approaches are often adopted. For example, Boosted Decision Trees \cite{boosted} and Feedforward Neural Networks \cite{Kolanoski1996} are much used in this context, but the latest state-of-the-art methods have not yet been fully explored and can bring a new light on the torrent of data being generated by experiments like those at the Large Hadron Collider (LHC) at CERN.
 
In a first approach we have tested the use of convolutional networks for the classification of collisions at LHC using Open Data Monte Carlo samples. The Compact Muon Solenoid (CMS) experiment \cite{cms}  has been pioneer in the context of the LHC in making public the collision data collected by the detector, opening them to the international community in order to carry out new analyses or to use them for training activities. CMS Open Data is available from the CERN Open Data portal\footnote{\url{http://opendata.cern.ch}} and we also have a dedicated portal developed in our center\footnote{\url{http://cmsopendata.ifca.es}}. 

In order to apply deep learning techniques conceived for image classification, to the analysis of these collisions, we propose an innovative visual representation of the different physics observables. We train a convolutional neural network on these images, representing simulated proton-proton collisions, to try to distinguish a particular physics process of interest. In our example, we try to discriminate the production of a pair of quarks top anti-top (ttbar) from other processes (background).

\section{Deep Learning Techniques for Image Classification}

\subsection{Deep Learning Architecture}

The technique of image classification using CNNs is included in the scope of deep learning.  Deep learning is part of a broader family of machine learning methods based on learning data representations, as opposed to task specific algorithms. The performance of these processes depends heavily on the representation of the data and on the algorithm used \cite{representation_learning}.

Following previous successful work in other fields within our group (like plant identification \cite{IHeredia}), we have selected as CNN architecture the Residual Network model \cite{ResNet} (ResNet) who won the ImageNet Large Scale Visual Recognition Challenge in 2015 \cite{ILSVRC15}. 

The architecture of the ResNet model used consists of a stack of similar (so-called residual) blocks, each block being in turn a stack of convolutional layers. The innovation of this architecture is that the output of a block is also connected with its own input through an identity mapping path. This alleviates the vanishing gradient problem, improving the gradient backward flow in the network and allowing to train much deeper networks. We choose our model to have 50 convolutional layers (aka. ResNet50).

As deep learning framework we use the Lasagne \cite{Lasagne} module built on top of Theano \cite{Theano1}\cite{Theano2}. We initialize the weights of the model with the pretrained weights on the ImageNet dataset provided in the Lasagne Model Zoo. We train the model for 40 epochs on different top performing GPUs using Adam \cite{Adam} as learning rule. During training we apply standard data augmentation (as sheer, translation, mirror, etc), and after applying the transformations we downscale the image to the ResNet standard input size (224\texttimes 224 pixels)\footnote{Code available at \url{https://github.com/IgnacioHeredia/plant_classification}}.

The preprocessing of the samples and the image generation has been done in Python\footnote{Code available at \url{https://github.com/CeliaFernandez/Image-Creation}}. The images have been generated extracting the simulated collisions data from a dedicated JSON file containing the main information on the physics observables at play. The JSON has been produced using a C++ framework\footnote{Code available at \url{https://github.com/laramaktub/json-collisions}} based on a template provided by the Open Data group to which the JSON generation part has been added. An example of the JSON file format used (\texttt{short.json}) together with the instructions to run the code are also found in the repository.

For didactic purposes, we describe in what follows some of the details of the image classification process.

\subsection{Image classification process}

The overall pipeline in CNNs is similar to standard NNs except for the fact that in this case we feed an image represented by a 3 dimensional tensor of shape 224\texttimes 224\texttimes 3 (image height\texttimes image width\texttimes RGB value). As in most machine learning algorithms, in this workflow we divide the image data in three splits (train\textbar val\textbar test) with roughly (80\textbar 10\textbar10) \% of the images. As always their respective roles are:

\renewcommand{\labelitemi}{$\bullet$}
\begin{itemize}
\item \textbf{Training set} \\
The data on this set is used to tweak the parameters of the net through an optimization process described below. We define the duration of the training process by the number of times (\textit{epochs}) we visit the whole training set. It is important to have a balanced dataset (ie. having a comparable number of images in each class) so that the abundance of a class is not a determining factor when classifying new images (ie. always predicting the more abundant class).
\item \textbf{Val set} \\
These data are used only as a check of how the trained net would perform on unseen data but they are never used to actually compute the net weights. They are useful for checking that the optimization process is correctly generalizing, not overfitting, and to eventually try different hyperparameters on another iteration of the training process (for example vary the number of layers or the number of training epochs).
\item \textbf{Test set} \\
These are holdout data, that one should not use until the very end of the workflow to finally assess the performance of our net. Having these as a separate set from the val set makes sense so as to not overfit the val set by appropiately tweaking the hyperparameters.
\end{itemize}

\begin{figure}[t]
\centering
\includegraphics[width = 10cm]{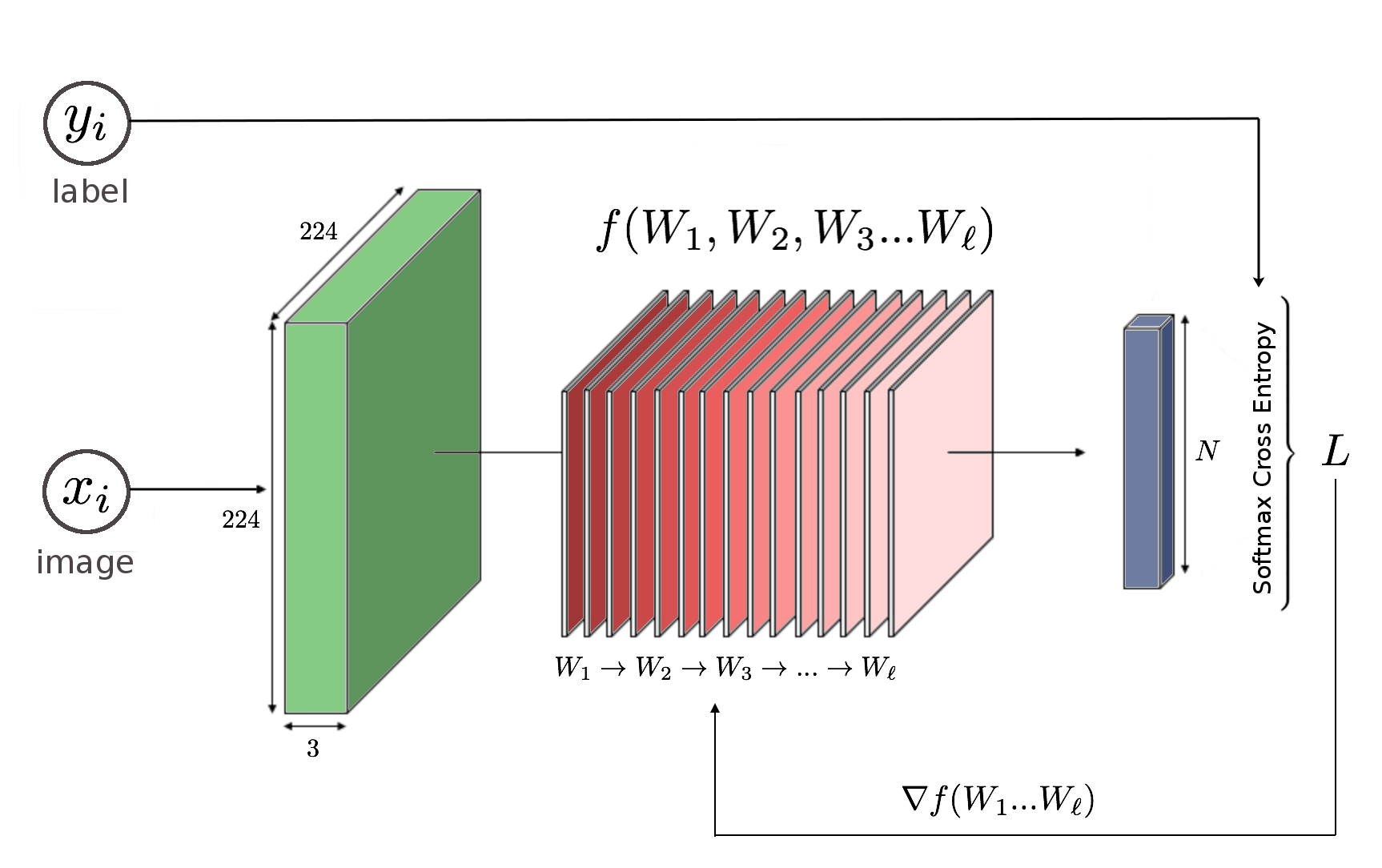}
\caption{Diagram of the optimization pipeline in the neural network. A given image $x_i$ (described as a tensor of shape 224\texttimes 224 \texttimes 3) of a known predefined class $y_i$ is given as an input to the net. Through a function $f(W_1, W_2... W_\ell)$ an $N$-dimensional vector is computed indicating the probability of the image to belong to one of the $N$ predefined classes. Using this vector and the true label $y_i$ a scalar value $L$ is generated. From there a backward signal is sent, gradually computing the gradients of the weights $W_i$ so as to minimize the value of the loss $L$.}
\label{FIG: Esquema CNN}
\end{figure}

Each optimization iteration performed during the training/learning phase consists on two steps:
\begin{itemize}
\item \textit{The forward pass} where we feed an image to the net and compute the upper layer (a vector of length $N$ where $N$ is the number of classes) using a \textit{score function} $f(W_1, W_2..., W_\ell)$ where $W_i$ are the \textit{weights} of the function. If we apply the softmax operation to this upper layer vector, each element of this vector can be seen as the probability of the image of belonging to that particular class. Using this vector and our knowledge of the correct class $y_i$ we can compute the \textit{loss}, which is a scalar value that measures how far the actual prediction is from the true label. The higher the loss, the worst the prediction, therefore the whole optimization process aims at finding set of optimal weights that lead to a minimum of the loss function. As in most classification problems, here we use the softmax cross-entropy as loss function.
\item \textit{The backward pass} where we compute the gradient of the loss with respect to the net's weights using the so called \textit{backpropagation algorithm} \cite{backpropagation}. Then we use the computed gradient to update the value of the weights according to a \textit{learning rule} (in our case Adam \cite{Adam}). Because the loss is averaged only over a batch of images, typically composed of tens of images (depending on your GPU memory), and not over the full dataset, we call this algorithm \textit{stochastic gradient descent}.
\end{itemize}
The whole pipeline is shown in Figure \ref{FIG: Esquema CNN}. Once the optimization process is done for a given number of epochs, we freeze the values of the weights and we are ready to perform inference with the network. Due to the fact that during training we have to perform both of these passes, most of the computational effort is carried here, while a test time we only have to perform the forward (inference) pass, being therefore much quicker.

\section{Representing Particle Collisions as Images}

The main innovation of this work is the way in which the collisions are represented as images. 
Collisions, also known as events, recorded in a HEP experiment by a detector like CMS \cite{CMS-Detector}, are described by a set of variables measured corresponding to the particles detected: the momentum of muons, electrons, photons and hadrons produced in the collision of the two accelerated protons, that are determined by the different subdetectors (tracking system, calorimeters, muon system, etc.). 
Along the global reconstruction of the event, new variables like the definition and momentum of jets are also introduced. 
The analysis of events uses these sets of variables to discriminate the events corresponding to the physics analysis channel of interest from the background.
So the most relevant observables in a collision correspond to the momenta (energy and direction) of the reconstructed particles, and also jets or other global variables, like the missing energy, in the event. 

As already stated, when generating the images for classification, the design of the event representation is crucial. All the observables are to be represented using a canvas of dimension 224\texttimes 224 pixels.

As explained below, in our approach each particle or physics object is represented as a circumference with a radius proportional to its energy, and centered in the canvas at a position corresponding to its momentum direction. The momentum direction use as coordinates the pseudorapidity $\eta$, related to the polar angle, and the azimuthal angle $\varphi$, which are standard choices in experiments with cylindrical symmetry. Additionally, we associate the color of the circumference to the type of particle or physics object represented. 

There are several considerations that have been take into account when proposing this representation, that are briefly discussed in what follows.

\subsection{Implementing the representation of physics objects}

We have considered the following points to define the transformation of the physics objects into their representation as an image: 

\begin{itemize}
\item \textbf{Resolution}\\
 Each physics object will be represented by a circumference with a radius defined as a function of its energy. As it is drawn using a discrete number of pixels, the scale must be chosen to accomodate the different ranges of energies while preserving as much as possible the resolution in energy. 

\item \textbf{Out of range representation}\\
When increasing the scale, the low energy objects can be better differentiated but circumferences corresponding to high energy objects could exceed the canvas size causing a misinterpretation. This is the main reason to discard a lineal dependency with the energy.

\item \textbf{Overlapping}\\
If the particles have relatively close $\eta$ and $\varphi$ values for their momenta directions, the corresponding representations may overlap. This is the main reason to chose  circumferences instead of full circles for their representation. One future direction could be looking at full circles with some transparency and see how it compares with the current approach.  
\end{itemize}

The use of a logarithmic scale to transform the energy of the physics object into a radius for the circumference representing it, allows us to reach a  balance between the previous factors:

\begin{equation}
R = C\cdot\ln\left(E\right)
\label{EQ: R}
\end{equation}\\

where the value $C$ is an effective scale factor that allows us to conciliate the previous points for the collisions being studied, providing the conversion into pixel units.\\

The center of the circumference, also in pixels units, is obtained using conversion factors $6/224$ along the $\eta$ axis and $2\pi/224$ along the $\varphi$ one, corresponding to the ranges $[-3,3]$ for $\eta$ and $[-\pi, \pi]$ for $\varphi$.

\begin{figure}[t]
\centering
\includegraphics[width=10cm]{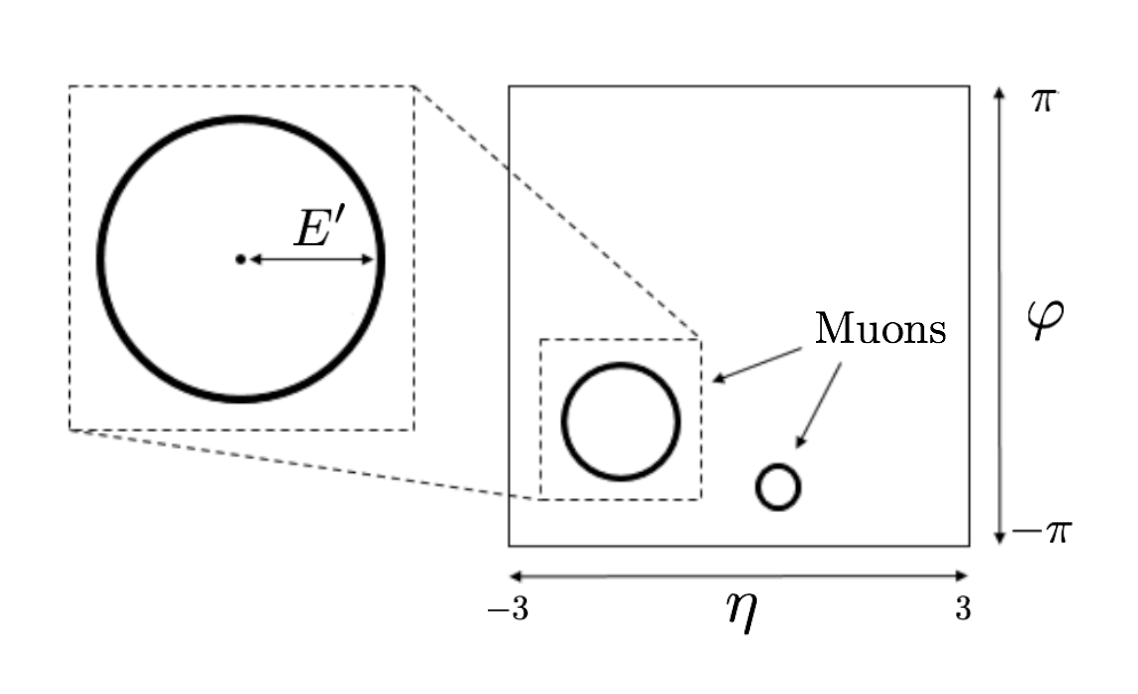}
\caption[]{Muon images diagrams. The muons are represented as circumferences with radius proportional to the logarithm of the energy. The horizontal position of the particles corresponds to the pseudorapidity $\eta$ within the range $\left[-3,\text{ }3\right]$. The vertical position shows the azimuthal angle $\varphi$ within the range $\left[-\pi, \text{ }\pi\right]$.}
\label{FIG:MuonImagesScheme}
\end{figure}

Figure \ref{FIG:MuonImagesScheme} presents a diagram of this representation for a single particle (a muon). More complex examples will be shown later.

\section{Physical variables: test on dimuon objects}

After chosing the previous representation, a first basic test was done using dimuon objects, to check if the neural network could separate different invariant masses patterns, a key feature in the reconstruction of collisions.\\

Anticipating the positive outcome of the test, we remind the direct relationship expected between the invariant mass of the dimuon object and the position and size of the circumferences used to represent them, as the invariant mass of a system with two particles 1 and 2, is\\
\begin{equation}
m^2 = m_a^2 = m_1^2 + m_2^2 + 2\left(E_1 E_2 - p_1 p_2 \cos \theta \right)
\end{equation}

or directly in terms of the variables used to define the representation of the particles as circumferences, the pseudorapidity $\eta$, the azimuthal angle direction $\varphi$ and the transverse momentum $p_T$ 
\begin{equation}
m^2 = 2p_{T1}p_{T2}\left(\cosh\left(\eta_1 - \eta_2\right) - \cos\left(\varphi_1 - \varphi_2\right)\right)
\end{equation}\\

\begin{figure}[t]
\centering
\subfigure[]{\label{SUBFIG:2011event_224}
\fbox{
\includegraphics[width=0.25\textwidth]{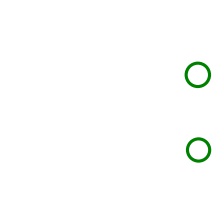}
}
}
\subfigure[]{\label{SUBFIG:2011event_46559}
\fbox{
\includegraphics[width=0.25\textwidth]{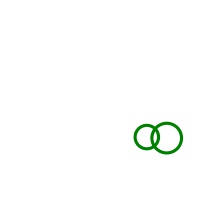}
}
}
\subfigure[]{\label{SUBFIG:2011event_122741}
\fbox{
\includegraphics[width=0.25\textwidth]{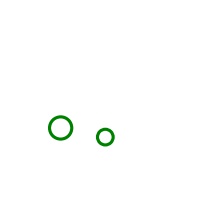}
}
}
\subfigure[]{\label{SUBFIG:2011event_275073}
\fbox{
\includegraphics[width=0.25\textwidth]{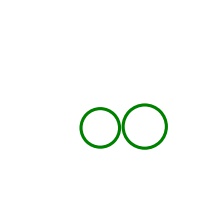}
}
}
\subfigure[]{\label{SUBFIG:2011event_96}
\fbox{
\includegraphics[width=0.25\textwidth]{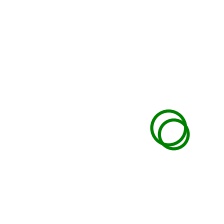}
}
}
\caption{Dimuon case. Examples of images, with different invariant masses $m$, belonging to \textbf{(a)} the \textit{None} class (with $m_{a} = 15,52$ GeV/c$^2$), \textbf{(b)} a $J/\Psi$ decay (with $m_{b} = 3,01$ GeV/c$^2$), \textbf{(c)} to a $\Psi'$ decay (with $m_{c} = 3,72$ GeV/c$^2$), \textbf{(d)} a $\Upsilon$ decay (with $m_{d} = 9,80$ GeV/c$^2$) and \textbf{(e)} a $Z$ decay (with $m_{e} = 95,76$ GeV/c$^2$). The x-axis depicts the pseudorapidity $\eta$ while the y-axis depicts the azimuthal angle $\varphi$.}
\label{FIG: Imagenes eventos masa invariante}
\end{figure}

To test the performance of the CNN to discriminate dimuon objects with different invariant mass, we have selected a sample of such objects from real events extracted also from CMS Open Data. 
The criteria to define such samples is shown in Table \ref{TAB:CriteriosClaseMuones}. Figure \ref{SUBFIG:2011event_224}-\ref{SUBFIG:2011event_96} shows examples of images corresponding to these different dimuon classes. 

\begin{table}
\begin{center}
\label{TAB:CriteriosClaseMuones}       
\caption{Criteria used to define different dimuon samples. Each range 1-4 is centered on the mass of a dimuon object corresponding to a  dimuon resonance.}
\begin{tabular}{c c c}
\hline
\textbf{Class} & \textbf{Dimuon object} & \textbf{Invariant mass range} (GeV/c$^2$)\\
\hline
1 & $J/\Psi$ & $\left[2.94,\text{ } 3.24\right]$ \\
2 & $\Psi'$ & $\left[3.65,\text{ } 3.95\right]$ \\
3 & $\Upsilon$ & $\left[6.46,\text{ } 12.46\right]$ \\
4 & $Z$ & $\left[83.69,\text{ } 98.69\right]$ \\
0 & None &  All other mass ranges \\
\hline
\end{tabular}
\end{center}
\end{table}

\subsection{Results on the classification of dimuon objects}

The number of images used for training, validation and test for each class is indicated in Table \ref{TAB:DIMUON-SETS}. Some of the training images have been clonated to assure a fair balance among the five categories although classes were already roughly balanced. Note that we solve the class imbalance problem by simply replicating images instead of using some clever modification of the loss function, like the weighted cross entropy, where we could have given higher weights in the loss computation to classes with lower abundance, instead of revisiting them, thus leading to a faster training. This dummy approach was taken so as to be able to reuse the code from \cite{IHeredia} with the least amount of overhead.\\

\begin{table}
\begin{center}
\caption{Information about the different datasets used in the dimuon case. In the case of \textit{train set} the values are obtained after some minor image clonation to equally populate the different classes.}
\begin{tabular}{|c|c|c|c|c|c|c|}
\hline
\textbf{Class} & \multicolumn{2}{|c|}{\textbf{train set}} & \multicolumn{2}{|c|}{\textbf{val set}} & \multicolumn{2}{|c|}{\textbf{test set}}\\
\hline
\hline
0 & 46584 & $21\%$ & 4250 & $44\%$ & 4307 & $45\%$ \\
\hline
1 & 44016 & $20\%$ & 2175 & $23\%$ & 2107 & $22\%$ \\
\hline
2 & 46284 & $21\%$ & 127 & $1\%$ & 131 & $1\%$ \\
\hline
3 & 45514 & $21\%$ & 2993 & $31\%$ & 3016 & $31\%$ \\
\hline
4 & 38109 & $17\%$ & 129 & $1\%$ & 114 & $1\%$ \\
\hline
\hline
\textbf{Total} & \multicolumn{2}{|c|}{220507} & \multicolumn{2}{|c|}{9674} & \multicolumn{2}{|c|}{9675}\\
\hline

\end{tabular}
\end{center}
\label{TAB:DIMUON-SETS}
\end{table}

\begin{figure}
\centering
\subfigure[Training accuracy]{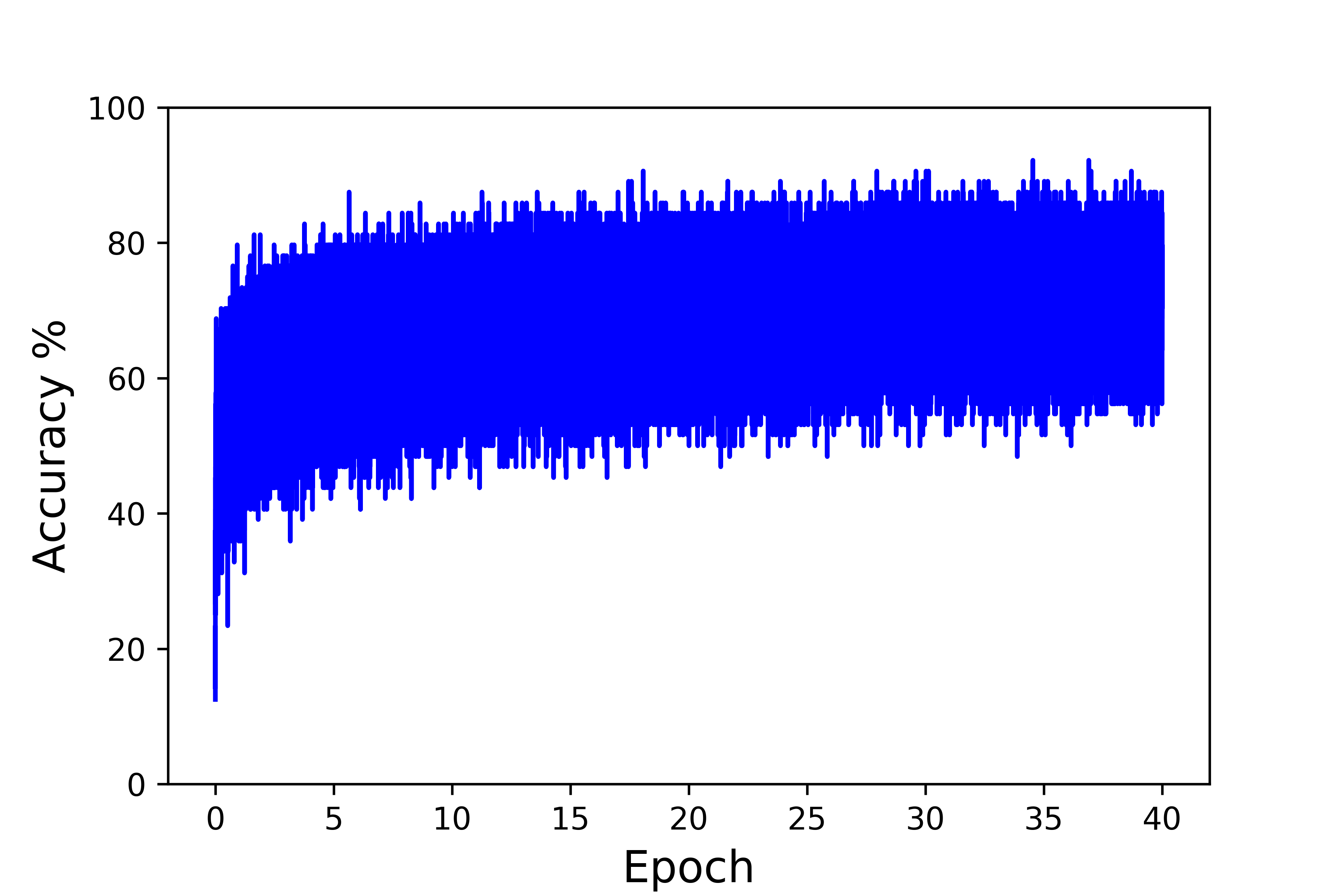
\label{SUBFIG: Muons_Accuracy.png}
\includegraphics[width=0.45\textwidth]{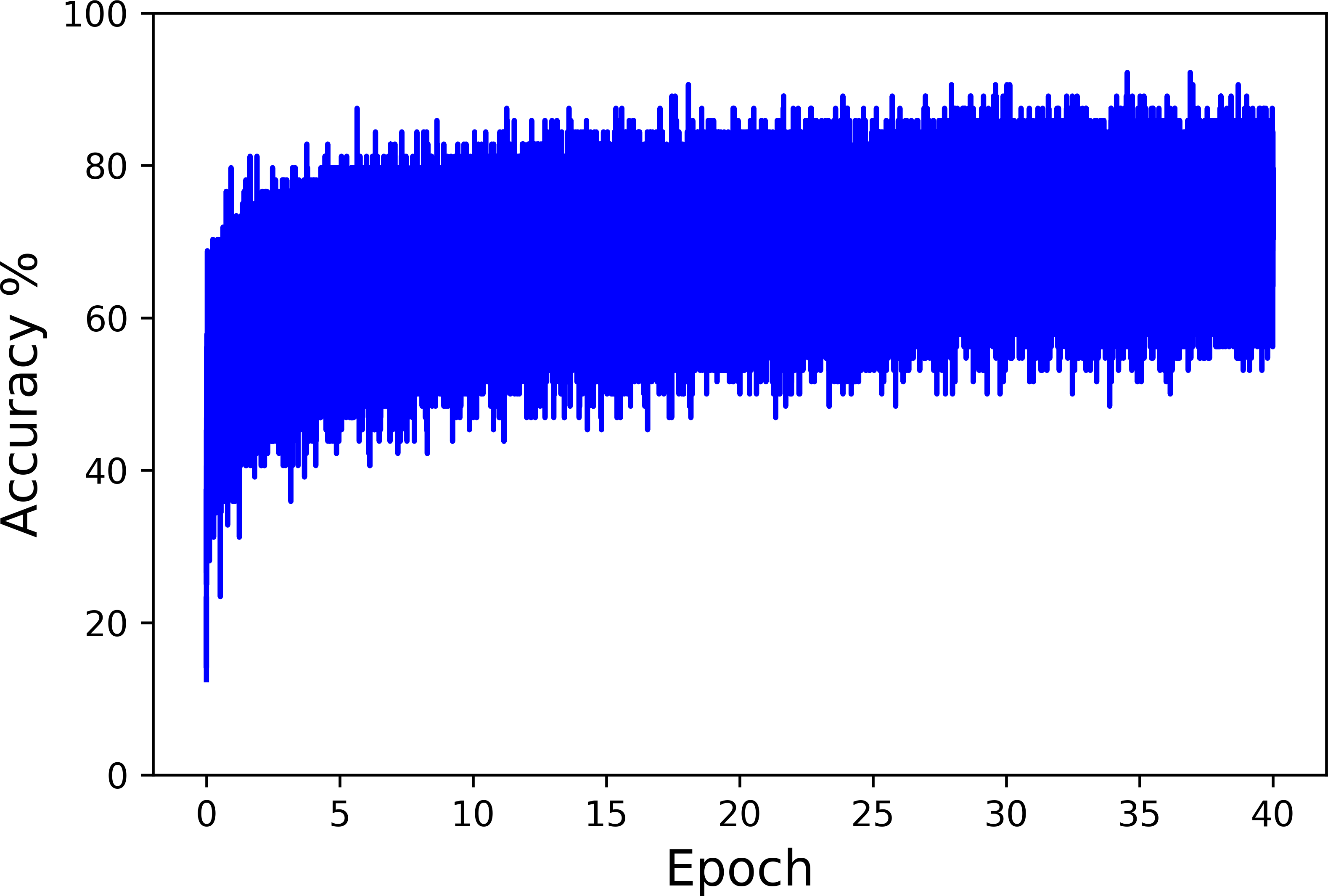}
}
\hspace{0.5cm}
\subfigure[Training loss]{\label{SUBFIG: Muons_Loss.png}
\includegraphics[width=0.45\textwidth]{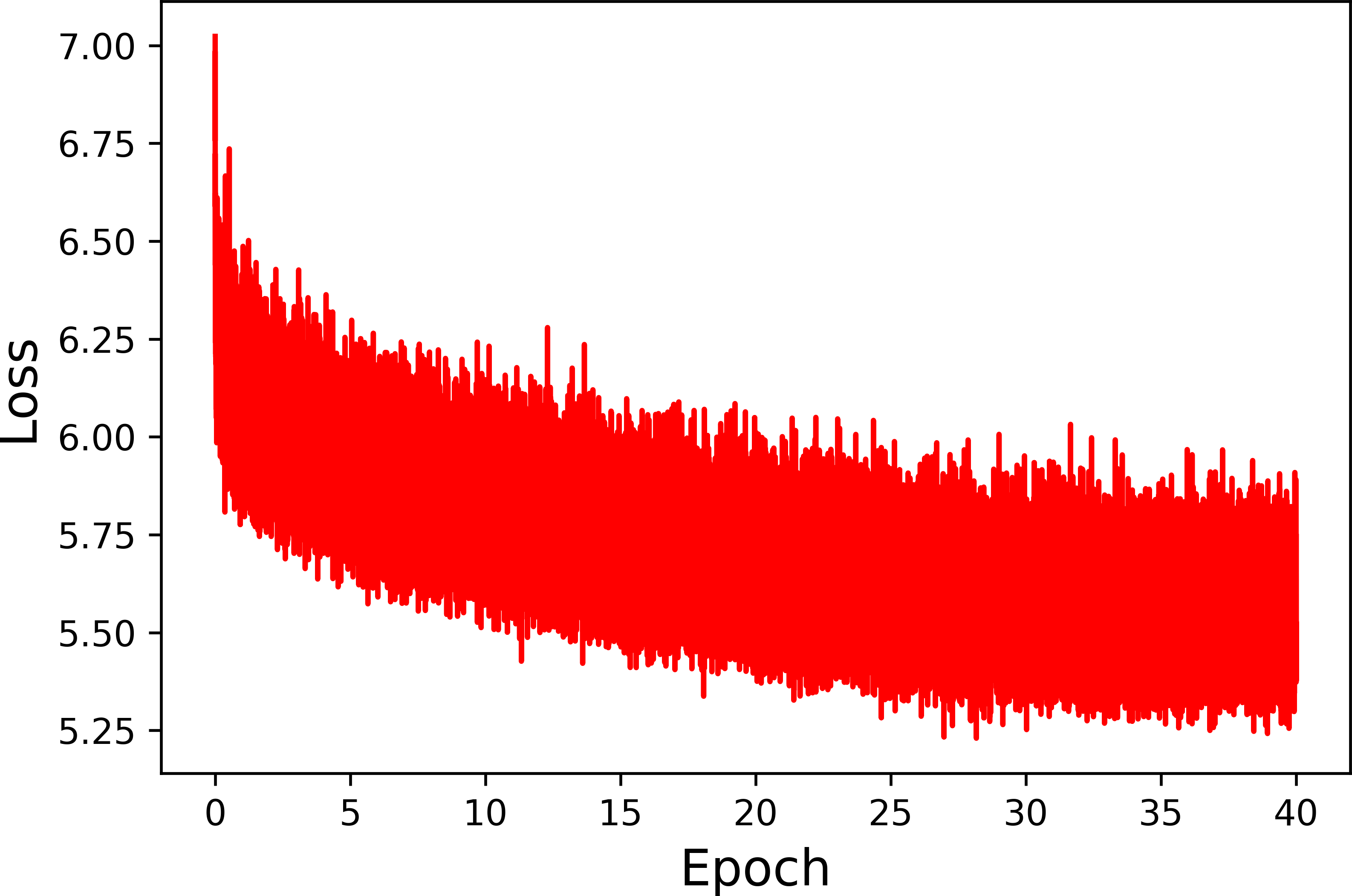}
}
\caption{Accuracy and loss training values for the dimuon case as a function of the number of epochs of the training.}\label{FIG: Precisión y perdida del entrenamiento con muones}
\end{figure}

We have set the training duration to 40 epochs and, as expected, the network accuracy increases as the loss discreases, eventually reaching a plateau, as can be seen in Figure \ref{FIG: Precisión y perdida del entrenamiento con muones}.\\

\begin{figure}
\centering
\subfigure[Non-normalized confusion matrix]{\label{SUBFIG: Muon-Non-Normalized_Confusion_Matrix.png}
\includegraphics[width=0.45\textwidth]{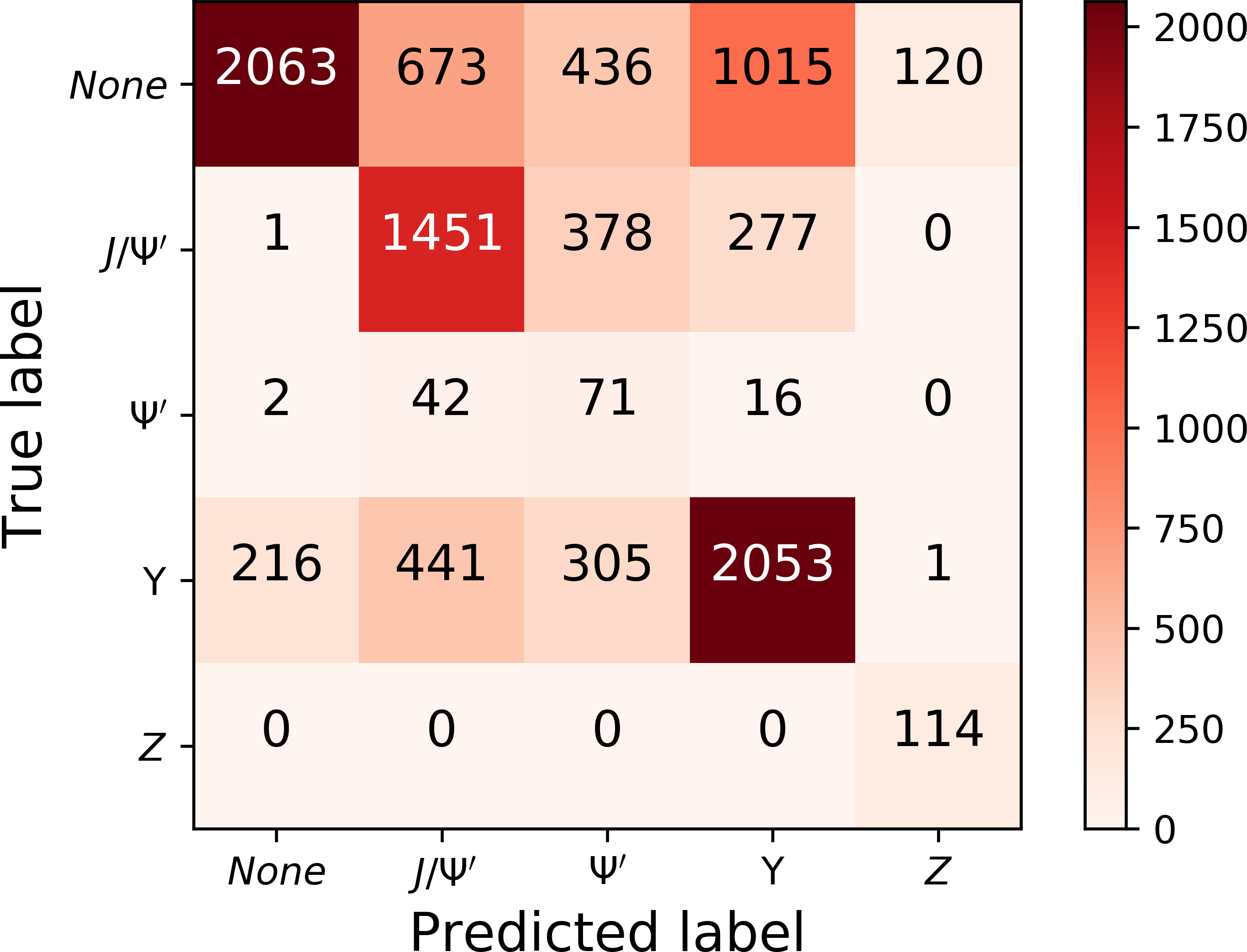}
}
\hspace{0.5cm}
\subfigure[Normalized confusion matrix]{\label{SUBFIG: Muon-Normalized_Confusion_Matrix.png}
\includegraphics[width=0.45\textwidth]{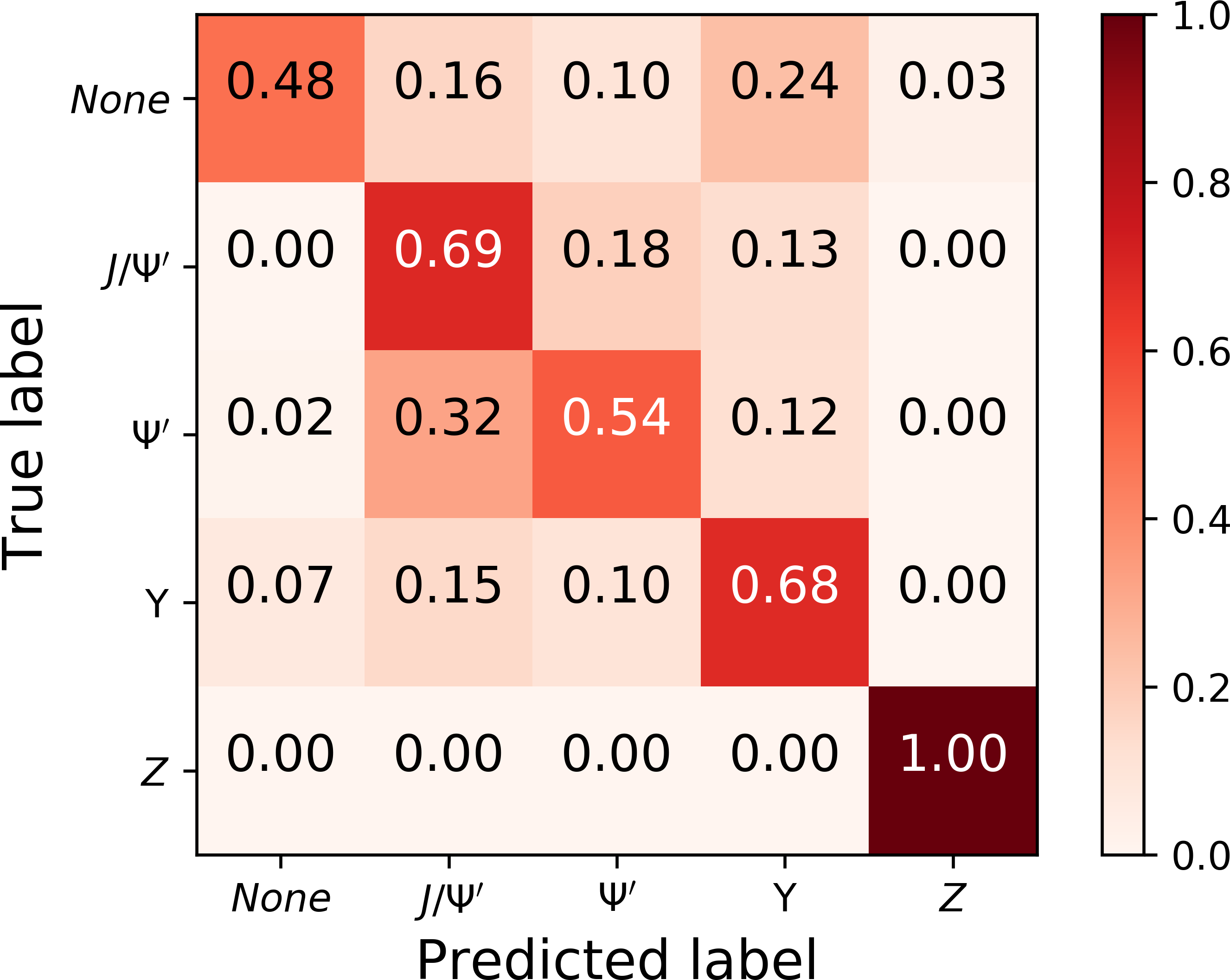}
}
\caption{Confusion matrices for the test set in the dimuon case using convolutional neural networks.}\label{FIG: Matrices de confusión Muones}
\end{figure}

Once the NN weights have been determined, the performance of the network has been evaluated on the test set. Figure \ref{FIG: Matrices de confusión Muones} show the corresponding the confusion matrix. As it can be seen, the network is capable of distinguishing quite well the images corrresponding to dimuon objects with an invariant mass close to the $Z$ boson mass, while the performance to discriminate dimuon objects with similar and lower invariant masses decreases significantly.

\section{Application to Complex Events}

After the encouraging initial test on dimuon objects, we have addressed a second exercise, using events corresponding to simulated collisions at 7 TeV at LHC recorded by the CMS detector \cite{CMS-Detector}, that have been released as Open Data by the CMS collaboration.

We have chosen as physics channel the production of top quark pair events, where each top quark decays into a W boson and a bottom quark. We want to select collisions where one of the W bosons decays leptonically into a charged lepton, electron or muon, with an associated neutrino. Although complex, these events provide a clear experimental signature, with an isolated lepton with high-transverse momentum, hadronic jets and a large missing transverse energy. We have considered as background processes the production of events where a W boson is produced in association with additional jets ($W+jets$ events) and events corresponding to the so called \textit{Drell-Yan} processes. The CMS publication webpage\footnote{\url{http://cms-results.web.cern.ch/cms-results/public-results/publications/TOP/7TEV.html}} on top physics results at 7 TeV provides a description of the interest of this physics analysis channel and detailed presentations of the involved processes, methods and results. 
All three samples \cite{dataset-DY}\cite{dataset-Wjets}\cite{dataset-tt} are obtained from the CMS Open Data portal.

\subsection{Event selection}

Before starting the learning, we need to make a preselection of the events according to the physics channel of interest.

We will focus on events having one lepton with a transverse momentum greater than 20 GeV fulfilling all the standard quality criteria for isolation and identification.
We select jets with a transverse momentum, $p_T$, greater than 30 GeV and within the angular range defined by $|\eta|$ $<$ 2.4.
We apply a b quark tagging discriminant (b-tagging), allowing us to identify (or "tag") jets originating from bottom quarks, by using the Combined Secondary Vertex (CSV) which is based on several topological and kinematical secondary vertex related variables as well as information from track impact parameters. We also use and represent in the event images the Missing Transverse Energy (MET). 

\subsection{Image representation}
\label{ssec:compleximages}
Leptons and jets are represented as circumferences centered according to their values of $\eta$ and $\varphi$ and whith a radius proportional to their transverse momentum, scaled according to the expression 

\begin{equation}
p_T' = C\cdot\ln\left(p_T\right) 
\label{EQ: Pt'}
\end{equation}

where again $C \sim 10.5$ is the scale factor allowing all the elements to be represented within the 224\texttimes 224 pixels canvas. \\
Each type of particle and jet is drawn with a different color: blue for the electrons, green for the muons, light red for non-btagged jets and dark red for btagged jets. 

Additionally, the missing transverse energy is drawn as a black circumference in each collision, moving vertically (according to $\varphi_{MET}$), and horizontally centered at $\eta=0$. As before its radius scales logaritmically with the absolute value of the MET.

\begin{figure}[t]
\centering
\subfigure[\textit{Drell-Yan}]{\label{SUBFIG:DY_event_3150}
\fbox{
\includegraphics[width=0.25\textwidth]{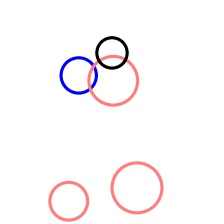}
}
}
\subfigure[$W+jets$]{\label{SUBFIG:Wjets_event_2690}
\fbox{
\includegraphics[width=0.25\textwidth]{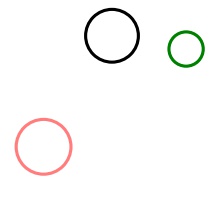}
}
}
\subfigure[$t\bar{t}+jets$]{\label{SUBFIG:TTjets_event_325119}
\fbox{
\includegraphics[width=0.25\textwidth]{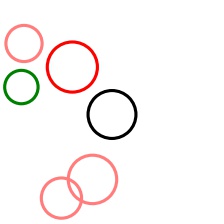}
}
}

\caption{Examples of images corresponding to the three different classes of collisions being classifed. The x-axis depicts the pseudorapidity $\eta$ while the y-axis depicts the azimuthal angle $\varphi$.}\label{FIG: Imagenes eventos MC}
\end{figure}

Figures \ref{SUBFIG:DY_event_3150}-\ref{SUBFIG:TTjets_event_325119} show sample images corresponding to the different classes of events under study. \\

\subsection{Results using CNN for classification of complex events}

The objective is to be able to differentiate between $t\bar{t}+jets$ events, and those corresponding to \textit{Drell-Yan} and $W+jets$ processes. 

The CNN is trained using Monte Carlo samples from CMS Open Data, with the statistics indicated in the Table \ref{TAB:TRAINING-TTBAR}. As done  previously, we clone some training images to enforce class balance.  

	\begin{table}

	\caption{Set of images for \textit{train set}, \textit{vat set} and \textit{test set}.  In the case of the \textit{train set} the information shown corresponds to the values before and after the image cloning for enforcing class balance. }
	\begin{center}
			\begin{tabular}{|c|c|c|c|c|c|c|c|c|}
				\hline
				\textbf{Clase} & \multicolumn{2}{|c|}{\textbf{train set} before} &\multicolumn{2}{|c|}{\textbf{train set} after} & \multicolumn{2}{|c|}{\textbf{val set}} & \multicolumn{2}{|c|}{\textbf{test set}}\\
				\hline
				\hline
				$t\bar{t}+jets$ & 30809 & $41,94\%$ & 30809 & $33,93\%$ & 5000 & $33,33\%$ & 5000 & $33,33\%$ \\
				\hline
				\textit{Drell-Yan} & 21709 & $29,54\%$ & 30000 & $33,04\%$ & 5000 & $33,33\%$ & 5000 & $33,33\%$ \\
				\hline
				$W+jets$ & 20950 & $28,52\%$ & 30000 & $33,04\%$ &  5000 & $33,33\%$ & 5000 & $33,33\%$ \\
				\hline
				\hline
				\textbf{Total} & \multicolumn{2}{|c|}{73468} & \multicolumn{2}{|c|}{90809} & \multicolumn{2}{|c|}{15000} & \multicolumn{2}{|c|}{15000}\\
				\hline
			\end{tabular}
		\end{center}
		\label{TAB:TRAINING-TTBAR}
	\end{table}

\begin{figure}
	\centering
	\subfigure[Non-normalized confusion matrix]{\label{SUBFIG: ttbarclassification-Non-Normalized_Confusion_Matrix.png}
		\includegraphics[width=0.45\textwidth]{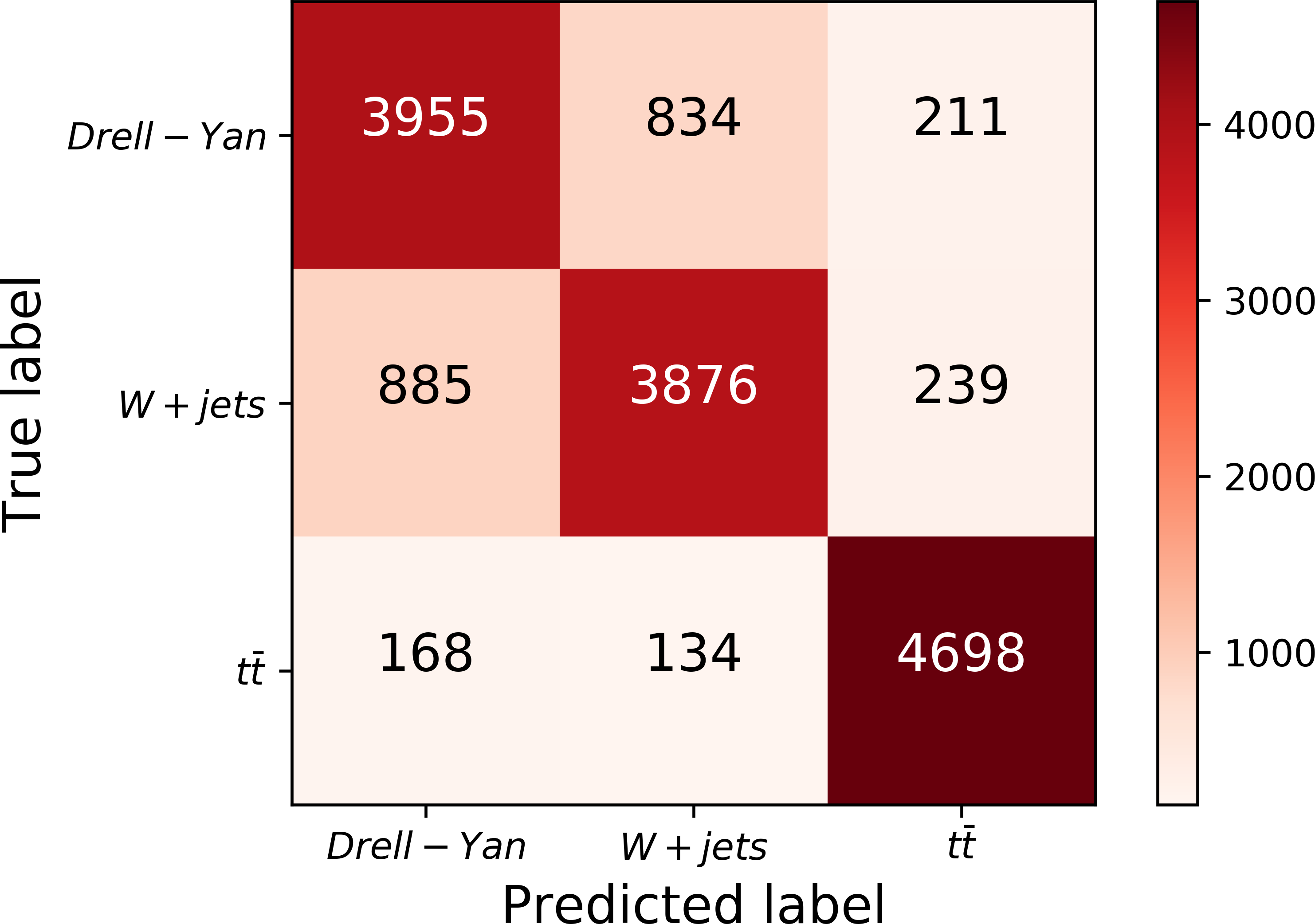}
	}
	\hspace{0.5cm}
	\subfigure[Normalized confusion matrix]{\label{SUBFIG: ttbarclassification-Normalized_Confusion_Matrix.png}
		\includegraphics[width=0.45\textwidth]{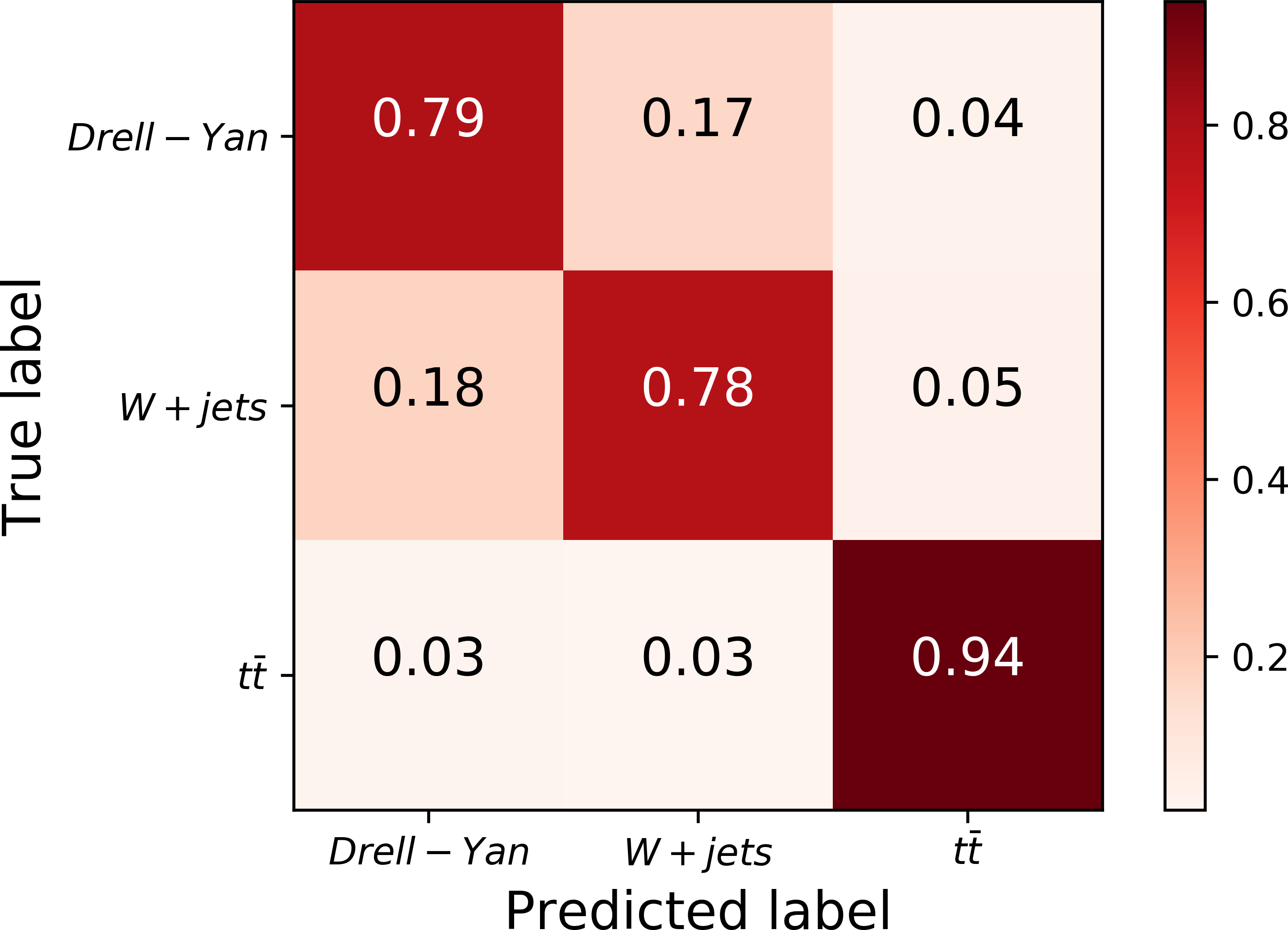}
	}
	\caption{Confusion matrices for the test set in the $t\bar{t}$ signal case using convolutional neural networks.}\label{FIG: Matrices de confusión ttbarclassification}
\end{figure}

The confusion matrix for the test set is shown in Figure \ref{FIG: Matrices de confusión ttbarclassification}. 
Approximately 94\% of the preselected ttbar events are correctly classified, while around 5\% of the W+jets and 4\% of the Drell-Yan events are incorrectly tagged as ttbar. In a signal ($t\bar{t}+jets$) and background (\textit{Drell-Yan} and $W+jets$) context, with 50/50 splits, the signal vs background discrimation efficiency would be 95,4\%.\\
We have also tried training the network defining only those two categories, signal ($t\bar{t}+jets$) and background (\textit{Drell-Yan} and $W+jets$). However it results in a slightly worse classification performance with a signalvs background efficiency of 93,6\%.

  If we want to use the CNN outputs as relevant variables in a physics analysis, the separation of the different background sources will likely result in a better control of systematic uncertainties.

\section{Comparison with a Feedforward neural network}

The results presented before have been compared with those obtained by using a simpler,  more direct, approach like deep \textit{feedforward} neural networks (FNNs). Recent work has already successfully applied many ideas of the deep learning community to the HEP field \cite{Baldi2014}.

Here we use a net of 5 hidden layers with 500 units per layers and standard 50\% dropout \cite{dropout} between layers. 

\begin{figure}
	\centering
	\subfigure[Non-normalized confusion matrix]{\label{SUBFIG: Muon-Non-Normalized_FeedForward.jpg}
		\includegraphics[width=0.45\textwidth]{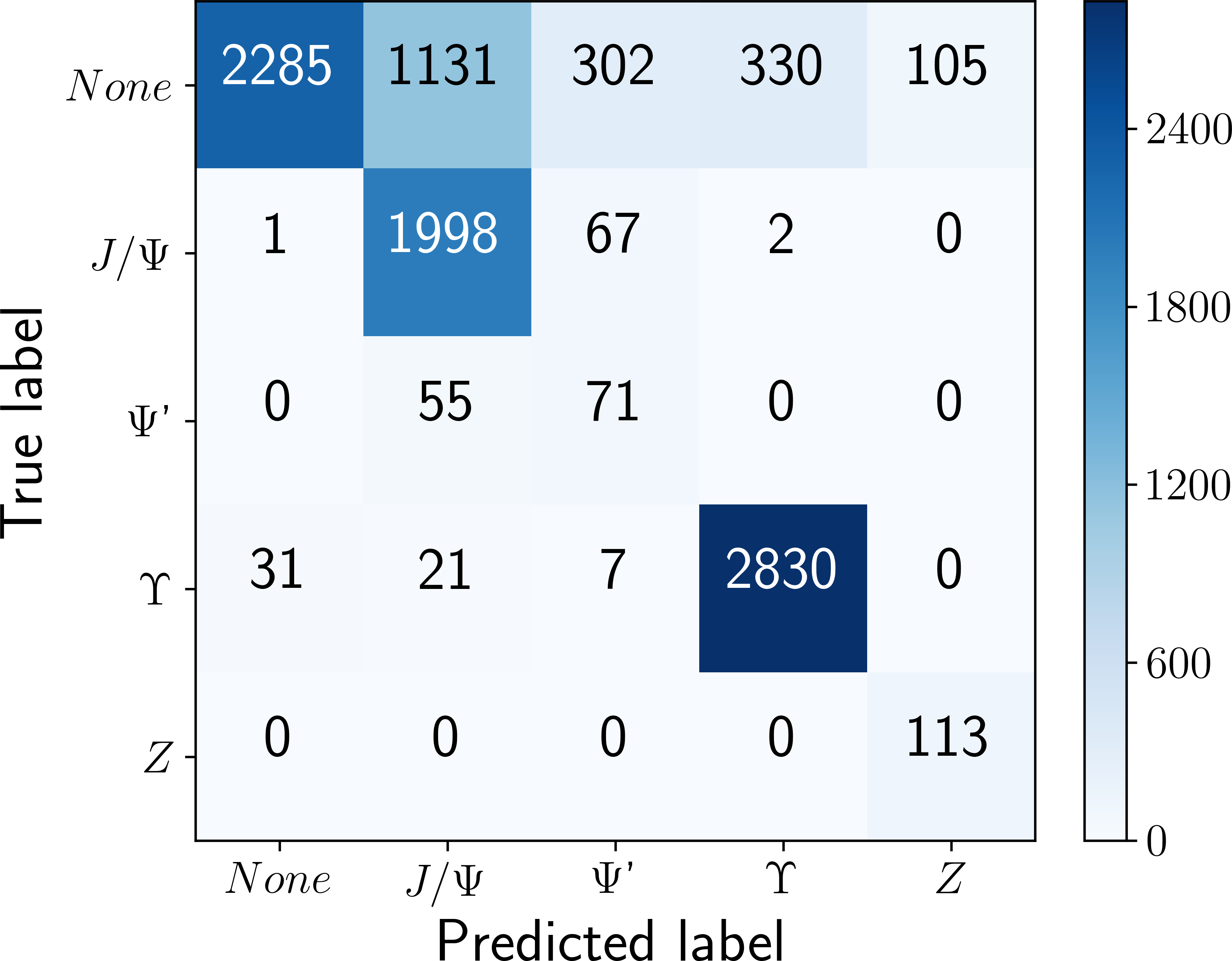}
	}
	\hspace{0.5cm}
	\subfigure[Normalized confusion matrix]{\label{SUBFIG: Muon-Normalized_FeedForward.jpg}
		\includegraphics[width=0.45\textwidth]{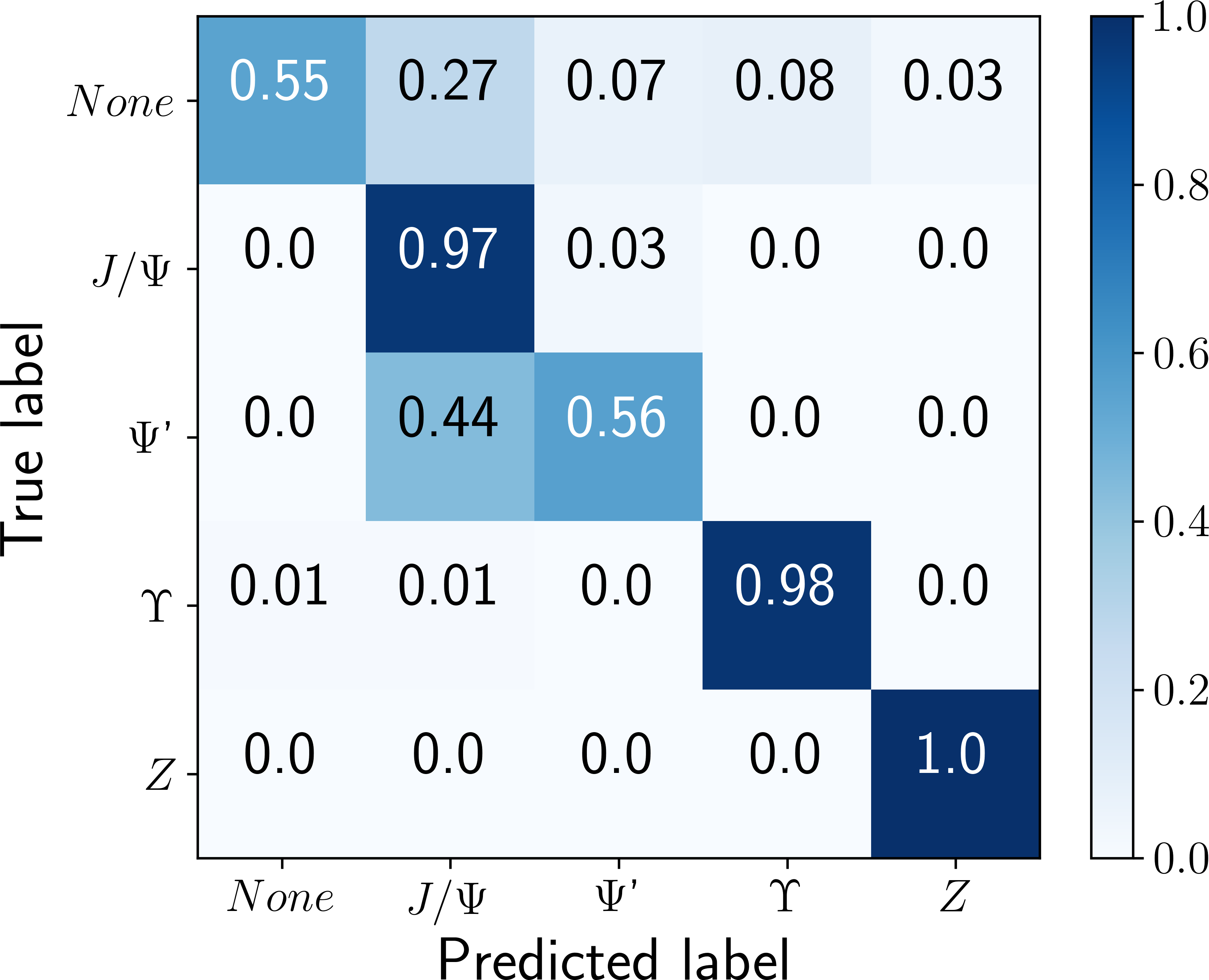}
    }
	\caption{Confusion matrices for the test set in the dimuon case using feedforward neural networks.}\label{FIG: Matrices de confusión Muones FeedForward}
\end{figure}

In the case of the classification of dimuon events according to their invariant mass results are shown in Figure \ref{FIG: Matrices de confusión Muones FeedForward}, where we can see, comparing to Figure \ref{FIG: Matrices de confusión Muones}, that FFNs are much more efficient at classifying all types of events, except for the \textit{None} class.\\

\begin{figure}
	\centering
	\subfigure[Non-normalized confusion matrix]{\label{SUBFIG: ttbar-Non-Normalized_FeedForward.jpg}
		\includegraphics[width=0.45\textwidth]{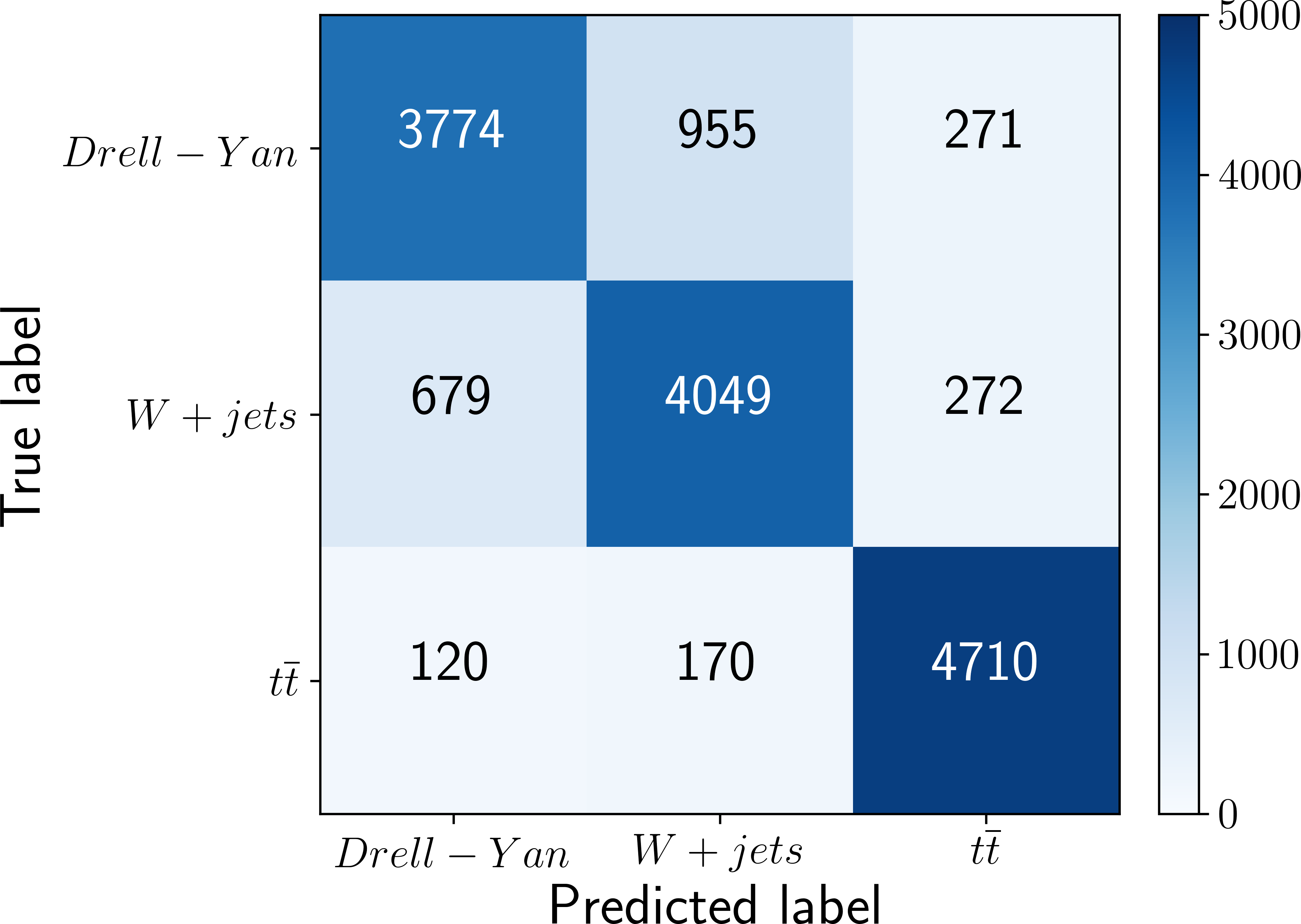}
	}
	\hspace{0.5cm}
	\subfigure[Normalized-confusion matrix]{\label{SUBFIG: ttbar-Normalized_FeedForward.jpg}
		\includegraphics[width=0.45\textwidth]{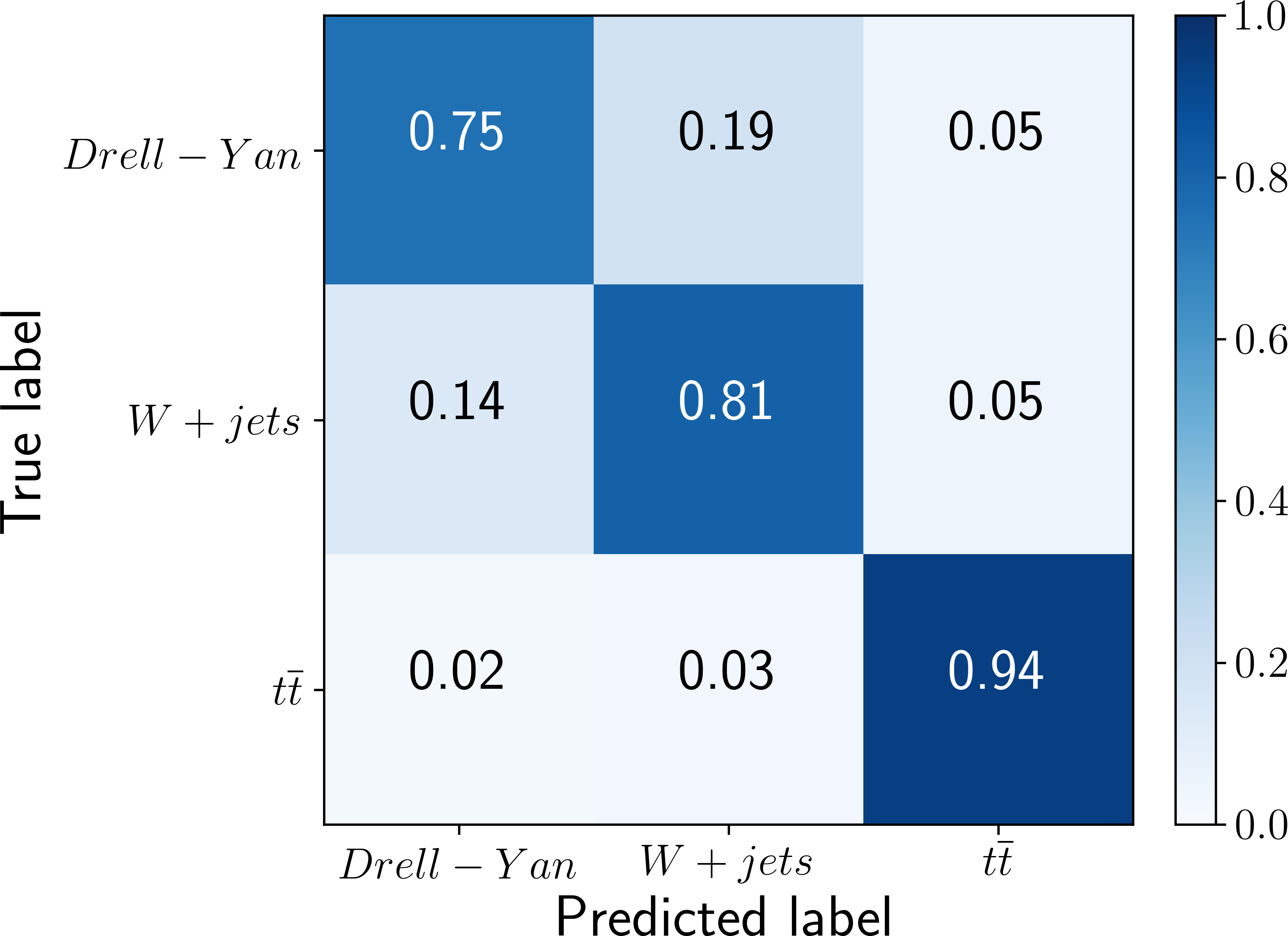}
	}
	\caption{Confusion matrices for the test set in the $t\bar{t}$ signal case using feedforward neural networks.}\label{FIG: Matrices de confusión ttbar FeedForward}
\end{figure}

In the case of $t\bar{t}$ vs background classification results are shown in Figure \ref{FIG: Matrices de confusión ttbar FeedForward}. In this case we can see, comparing to Figure \ref{FIG: Matrices de confusión ttbarclassification}, that FFNs are better at classifying $t\bar{t}+jets$ and $W+jets$ (but not \textit{Drell-Yan}). However, more importantly, we can see that CNNs would outperform FFNs in the signal vs background metric, with a  94,6\% efficiency for FFNs.\\

The advantages of FFNs compared to CNNs are that the preprocessing time is much shorter (as you only have to prepare a scalar vector of the variables instead of a full 224\texttimes 224\texttimes 3 tensor image) and that the training time is much faster (as they are shallower and the computation in between layers is usually much lighter).

The downside of FNNs is their vector representation of variables, which makes handling heteregenous (non fixed-size) data not very intuitive. In this case we handled the various length events by filling the empty parameters with default values. In contrast, in the CNN case adding one more particle to the event just implies drawing one more circle in the image.

\section{Conclusions}
\label{sec:conclusions}

The preliminary results presented in this study show that the use of Convolutional Neural Networks could be a promising tool to classify collisions in particle physics analysis.

An intuitive visual representation of the events has been proposed that enables the inclusion of the main observables used in high energy physics analysis into an image.\\

An initial test has shown that using this representation for dimuon objects, a CNN is able to classify them according to their invariant mass. 
\\ 
A second test has been applied to the classification of more complex events, using Open Data describing simulated collisions at LHC at 7 TeV in the CMS detector, and corresponding to three different physics processes, \textit{Drell-Yan}, $W+jets$ and $t\bar{t}+jets$. The test has returned promising initial results, correctly tagging signal and background events with an efficiency around 95\%, and comparing slightly favourably with other more direct methods, like standard feedforward NNs.\\

We plan to extend this work in the future to analyse, among other possibilities, its applicability to the classification of real data, having in mind the problems related to the uncomplete description usually provided by the simulation.

\section{Acknowledgements}
\label{sec:acknowledgements}

We would like to show our gratitude to the Data Preservation and Open Access group in CMS for all the valuable support and for providing insight and expertise that greatly assisted the realisation of this work. Ignacio Heredia is funded by the EU Youth Guarantee Initiative (Ministerio de Economia, Industria y Competitividad, Secretaria de Estado de Investigacion, Desarrollo e Innovacion, through the Universidad de Cantabria). Celia Fern\'andez is funded by a collaboration fellowship (Ministerio de Educaci\'on, Cultura y Deporte, through the Universidad de Cantabria).

\end{document}